\documentclass[letterpaper, 10 pt, conference]{ieeeconf}  
\IEEEoverridecommandlockouts                        
\usepackage[hyphens]{url}  
\usepackage{graphicx} 
\usepackage{amsmath, amssymb, amsfonts}
\usepackage{algorithm}
\usepackage{algorithmic}
\usepackage{placeins}   
\usepackage{graphicx}
\usepackage{hyperref}
\usepackage{fontawesome5}
\usepackage{titletoc}     
\usepackage{xcolor}
\usepackage{booktabs}
\usepackage[table]{xcolor}

\usepackage{etoc}
\usepackage{etoc}

\usepackage{fvextra}
\usepackage{multirow}
\definecolor{oursrow}{RGB}{232,240,254}

\usepackage{subcaption}  
\usepackage[dvipsnames]{xcolor}
\definecolor{darkgreen}{rgb}{0,0.5,0}
 
 \newcommand{\yuxuan}[1]{{\color{black} #1}}
  \newcommand{\hussein}[1]{{\color{black} #1}}
 \newcommand{\ihabnew}[1]{{\color{black} #1}}
 
\title{\LARGE \bf
CrossSafe: Towards Cross-Embodiment Latent Safety Filters
}

\author{
Ihab Tabbara$^{*}$,
Yuxuan Yang$^{*}$,
and Hussein Sibai%
\thanks{$^{*}$Equal contribution.}%
\thanks{The authors are with the Department of Computer Science and Engineering,
Washington University in St. Louis, St. Louis, MO, USA.
{\tt\small \{i.k.tabbara, y.yuxuan, sibai\}@wustl.edu}}
}

\begin{document}
\maketitle
\thispagestyle{empty}
\pagestyle{empty}

\begin{abstract}
Cross-embodiment learning has shown that a single model, such as a vision-language-action (VLA) model, can learn state representations and manipulation skills that can be applied across heterogeneous robots to accomplish various tasks. We hypothesize that the same holds for safety enforcement. The reasoning required to satisfy a safety constraint, such as detecting an obstacle, recognizing that it should be avoided, and selecting a safe abstract action, is largely shared across robots. What differs across embodiments is how the abstract safe action is realized: morphology, kinematics, and dynamics determine which actions are safe and feasible. Consequently, the same action can be safe for one robot and unsafe for another. This is especially important for generalist manipulation policies that operate in a common end-effector action space without explicitly capturing how safety depends on the robot’s morphology and kinematics.  We propose  embodiment-conditioned safety filtering, in which a Hamilton--Jacobi reachability-based value function and its corresponding safety-maximizing policy are shared across robots. Using a morphology-aware latent representation of the robot and its environment, we perform Hamilton--Jacobi reachability analysis directly in latent space so that the learned safety concepts can generalize across embodiments while remaining explicitly conditioned on each robot's morphology and kinematics. We evaluate our approach across five bimanual robot embodiments and five manipulation tasks with whole-body collision-avoidance constraints. Our results show that a single policy, jointly trained across five manipulation tasks and four embodiments, exhibits zero-shot generalization to a held-out embodiment, reducing the nominal policy's collision rate. They also show that training using more embodiments improves generalization. Videos and code are available at \url{https://trustworthyautonomy.github.io/CrossSafe/}
\end{abstract}

\begin{figure*}[t]
    \centering
    \includegraphics[width=1\textwidth]{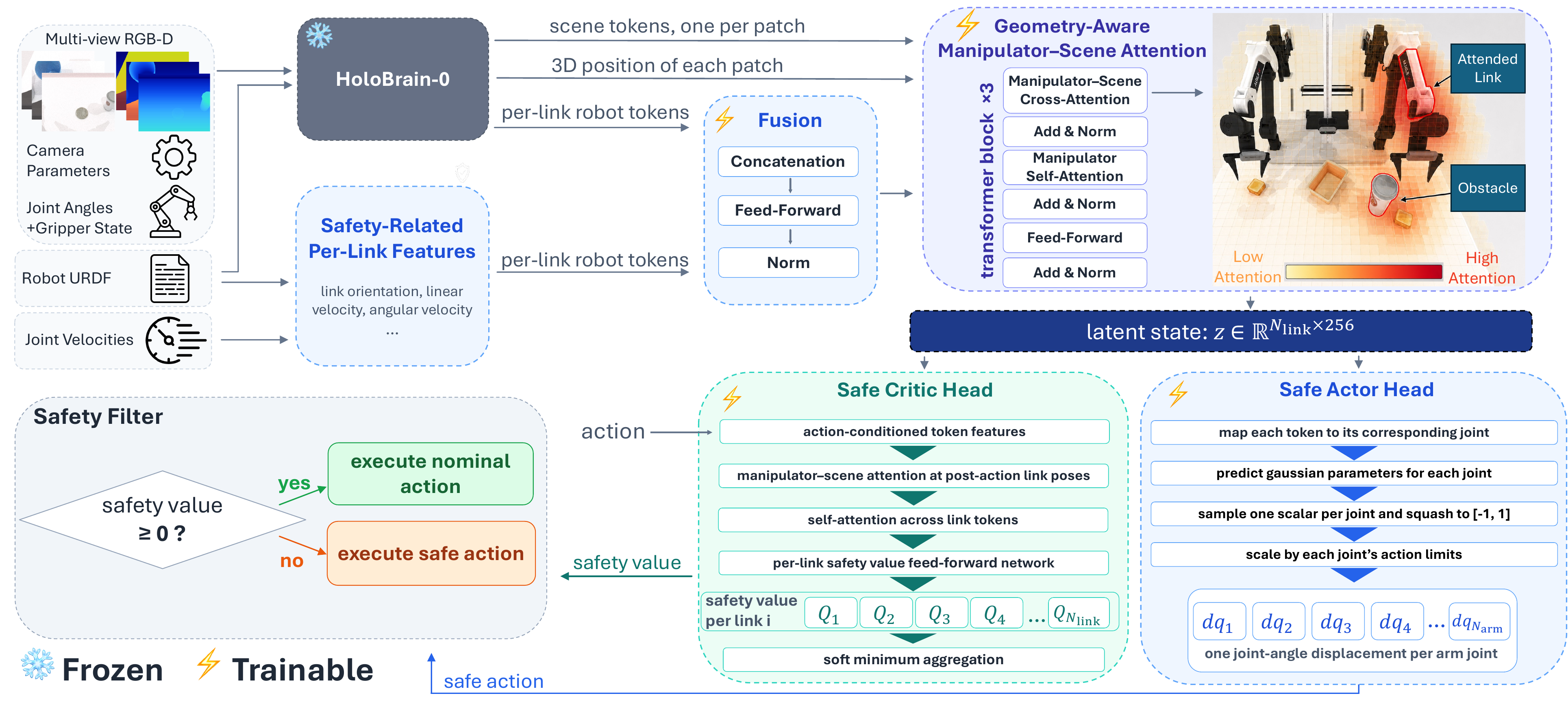}
\caption{\textbf{Overview of CrossSafe.} A frozen HoloBrain-0 produces scene and per-link robot tokens, which we augment with safety-related per-link features not encoded by HoloBrain-0. Geometry-aware cross-attention produces a safety-aware latent state with one token per robot link. The safety critic and actor use this latent state to estimate safety and output per-joint angle displacements, respectively, generalizing across embodiments with different degrees of freedom (DoFs).
}
    \label{fig:main}
\end{figure*}



\section{Introduction}

Scaling robot learning by collecting and training on separate datasets for every robot platform is costly and limits the reuse of experience across embodiments. Cross-embodiment learning addresses this problem by sharing task knowledge across robots with different morphologies and control interfaces. This idea has been explored across multi-robot policies \cite{openx2024,octo2024,crossformer2024}, morphology-aware manipulation models \cite{xvla2025,holobrain2026,aceego2026}, and cross-gripper or dexterous-hand generalization  \cite{graspgenx2026,dexgraspzero2026}. These results suggest that substantial task knowledge can be shared across physically different robots. We ask whether the same principle can be extended to safety filters: can data collected from multiple robot embodiments be used to learn a safety filter that generalizes to a novel one? 

Many generalist and cross-embodiment manipulation policies output actions in a common task-space representation across robots, such as delta end-effector pose commands \cite{octo2024,kim24openvla}, which can then be mapped to embodiment-specific joint motions. 
While this action representation lets a single policy's manipulation skills transfer across embodiments, the same end-effector trajectory can induce different whole-body motions on robots with different link lengths, joint axes, joint limits, and configurations.
Consequently, the same action may be safe for one embodiment but unsafe for another, and the safe control required to avoid entering the failure set may be embodiment-dependent.

This motivates a cross-embodiment state representation that preserves the safety-relevant information shared across robots while retaining sufficient information about the morphology and kinematics of each robot, together with an action space that \ihabnew{enables} different safe whole-body controls for different robots.

Existing cross-embodiment safeguarding methods address this differently: \ihabnew{EmbodiSteer \cite{embodisteer2026} corrects a given diffusion policy's actions at each denoising step using embodiment-specific kinematics and whole-body collision costs computed with cuRobo}, while Any-Body Guard \cite{anybodyguard2026} computes safe actions in each robot's native configuration space. Neither learns a  safety representation that supports jointly training a single safety value function and its corresponding safety-maximizing policy across heterogeneous robot embodiments. We therefore study 
embodiment-conditioned safety filtering, where a single Hamilton--Jacobi (HJ) value function is learned for 
heterogeneous robot embodiments. We seek a representation that captures safety-relevant information shared across robots while preserving morphology- and state-dependent information relevant for safety.

We build on HoloBrain-0 \cite{holobrain2026}, whose pretrained representation provides a natural starting point \hussein{as it jointly encodes} multi-view 3D scene information and variable robot kinematic structures using joint poses and graph-structured attention. We augment HoloBrain-0\hussein{'s representation} with safety-relevant per-link kinematic features and introduce geometry-aware manipulator-scene cross-attention that incorporates distances and directions between robot links and scene patches, expressed in a shared 3D coordinate frame. 
The resulting embodiment-conditioned  latent state allows training a shared safety critic to generalize across embodiments and assign different safety values to different embodiments. A corresponding policy can also be trained to produce embodiment-dependent safety-maximizing actions.

We evaluate our approach in RoboTwin 2.0 \cite{robotwin2} across five bimanual embodiments and five manipulation tasks under collision-avoidance constraints. Using five leave-one-embodiment-out folds, we jointly train the HJ value function and safe policy on four embodiments and evaluate them on the fifth, which is excluded from training of the safety-filter components. To our knowledge, CrossSafe is the first framework to jointly learn a single latent safety filter across multiple robot embodiments and demonstrate zero-shot generalization to embodiments excluded from safety-filter training.
Our contributions are: 
\begin{itemize}
    \item We introduce a Hamilton-Jacobi reachability-based  safety-maximizing actor and critic  whose latent state encodes a robot's proprioception, observations, and morphology, enabling generalization across embodiments, even ones with different degrees of freedom. 

    \item We evaluate CrossSafe on five bimanual robot embodiments and five tasks in RoboTwin~2.0, showing reduced collision rates on unseen embodiments without fine-tuning, and improved generalization when training on data from four embodiments instead of a single one.
\end{itemize}
\section{Related Work}

\subsection{Cross-embodiment generalist policies}
Large-scale multi-robot datasets and generalist policies such as Open X-Embodiment \cite{openx2024}, Octo \cite{octo2024}, OpenVLA \cite{kim24openvla}, $\pi_{0.5}$ \cite{pi0.5}, and CrossFormer \cite{crossformer2024} show the benefits of training across heterogeneous robot platforms. Recent cross-embodiment policies address embodiment variation through domain-specific prompts \cite{xvla2025}, morphology conditioning \cite{aceego2026}, shared action representations \cite{gearvla2026}, geometric interfaces \cite{cei2026}, end-effector traces \cite{bar2026}, or shared embodied reasoning \cite{zr02026}. Cross-embodiment generalization has also been studied for robot hands with heterogeneous end effectors \cite{graspgenx2026,dexgraspzero2026}.

A complementary line of work explicitly encodes the robot's morphology. HoloBrain-0 \cite{holobrain2026}, which we build on, combines multi-view 3D perception with URDF-derived kinematic priors, joint poses, and graph-structured attention over the kinematic chain. We extend this representation to also encode safety-critical features. Related methods encode morphology through kinematic graphs \cite{getzero2024}, a topology-aware end-effector graph paired with geometry-aware state tokens \cite{eagg2026}, 
morphology-agnostic encoder--decoder architectures for multi-embodiment locomotion \cite{onepolicy2024}, or morphology-conditioned world models \cite{qwm2026}.

Our goal is to generalize safety assurance across embodiments. In current VLA design practice, cross-embodiment pretraining is commonly followed by adaptation or fine-tuning to the target robot \cite{xvla2025,holobrain2026,kim24openvla,pi0.5,octo2024}. Recent methods such as LAP \cite{lap2026} and Cloak \cite{cloak2026} instead target zero-shot transfer by promoting embodiment-invariant task representations, but such invariance is less suitable for safety-critical control, where robot morphology  directly affects the set of possible safe actions. We therefore seek a state representation that generalizes across robots while preserving the embodiment-specific information needed for safe control.


\subsection{Latent safety filters and safe control}

Control barrier functions (CBFs)~\cite{ames2017cbf} and Hamilton--Jacobi (HJ) reachability~\cite{bansal2017hj} provide principled tools for synthesizing safety filters, while learning-based methods extend both frameworks to higher-dimensional systems, through neural CBFs~\cite{so2024train,tabbara2025learning} and through HJ methods that approximate the value function with reinforcement learning~\cite{fisac2019hjrl}.
Recent works extend safety filtering to learned latent representations \cite{nakamura2025latent,pvr_yuxuan_ihab}, 
neural operators \cite{li2025hjrno}, learned configuration-space barriers \cite{long2025neural}, reachability-based learned policies \cite{tayal2026safefql}, and temporal-logic specifications \cite{so2026value}. 
Language-conditioned HJ safety filters share one learned actor and critic across multiple language-specified safety constraints \cite{tabbara2026language}. A complementary line of work uses conformal prediction to bound the errors of
learned safety filters, recovering probabilistic safety guarantees~\cite{lin2024verification,tabbara2025statistically,seo2025unisafe,tayal2025cp,safe_control_sacha}. These advances generalize safety filtering across observations, constraints, environments, or specifications but do not address training a shared  safety value function that is explicitly conditioned on robot embodiments. 

More closely related to our work, EmbodiSteer \cite{embodisteer2026} corrects the diffusion policy's generated action at each denoising step using embodiment-specific robot kinematics and whole-body collision costs computed with cuRobo. \ihabnew{Any-Body Guard \cite{anybodyguard2026} certifies a local probabilistically safe polytope in each robot's configuration space, the space of joint angles $q \in \mathbb{R}^{N_\textit{DoF}}$ with one coordinate per joint, by sampling configurations and evaluating a violation function built from that robot's forward kinematics and an object-based scene representation. This is a sampling-based technique that does not train any models. It 
yields a probabilistic safety guarantee, but each robot's safe set is recomputed from its own kinematic model at runtime, so no safety reasoning is shared across embodiments. We instead amortize this into a safety value function learned offline jointly across robots. }

\section{Preliminaries}
\label{sec:prelim}
Hamilton–Jacobi (HJ) reachability is a formal control-theoretic framework for verifying control  systems' safety and synthesizing safe controllers~\cite{bansal2017hj}. 
Consider a dynamical system of the form $s_{t+1} = f(s_t, a_t)$, where $s_t \in S$ and $a_t \in A$ are the state and the action at time $t$, respectively. We denote the trajectory of the system starting from state $s$ and following a policy $\pi: S \rightarrow A$ by $\xi_{s}^\pi: \mathbb{N}^{\geq 0} \rightarrow S$.  Given a set of states $\mathcal{F} := \{s \mid h(s) < 0\}$ consisting of the {\em failure} (or {\em avoid}) states, where $h: S \to \mathbb{R}$ is a Lipschitz continuous function,
 HJ reachability analysis  computes the optimal  
 value function $V: S \to \mathbb{R}$ for {\em avoiding} $\mathcal{F}$, where $V(s) := \sup_{\pi}\inf_{t\geq 0} h(\xi_{s}^\pi(t))$. The latter  is the fixed point of the Bellman equation: $V(s) = \min \left\{ h(s), \max_{a \in A} V(f(s, a)) \right\}$. 
 
The associated optimal policy $\pi^*$ for {\em avoiding} $\mathcal{F}$ satisfies  $\forall s \in S, \pi^*(s) := \arg \max_{a \in A} V(f(s, a))$. The zero-sublevel set of $V$, i.e., the set $\{s\ |\ V(s) < 0\}$, is called the {\em backward reachable set} (BRS) of the system corresponding to  $\mathcal{F}$. It consists of the states starting from which the system will inevitably reach the failure set eventually under any policy, and thus is the largest set of {\em unsafe} states. 

Reinforcement learning-based approaches have been suggested to address the curse of dimensionality of HJ reachability
\cite{fisac2019hjrl}.
Actor-critic algorithms (e.g., SAC \cite{sac}) were used for systems with continuous action spaces. 
In such approaches, the parameters $\theta$ of the Q-function, which is also called the {\em HJ safety critic}, are optimized  by minimizing the loss function: 
\begin{equation}
L(\theta) := \mathbb{E}_{(s_t,a_t,s_{t+1})\sim \mathcal{D}}\left[(Q_\theta(s_t, a_t) - y_t)^2\right],
\label{eq:safety_q_loss}
\end{equation}
where $\mathcal{D}$ is the distribution of transitions $(s_t,a_t,s_{t+1})$ collected during environment rollouts
and the {\em target} $y_t$ is defined as follows: 
$
y_t := (1-\gamma)h(s_t) + \gamma \min\left\{h(s_t), \max_{a \in A} Q_{\theta}(s_{t+1}, a)\right\}$ with $\gamma\in(0,1)$~\cite{fisac2019hjrl}. 



\section{Morphology-Aware Latent Safety Filtering}
\label{sec:Morphology-Aware Latent Safety Filtering}
We learn a shared HJ safety critic $Q_\theta$ and a shared safe policy $\pi_{\phi}^{\mathrm{safe}}$ across embodiments, potentially with different numbers of links and joints. Both operate on a morphology-aware latent state produced by an encoder, which builds on the frozen, pretrained HoloBrain-0~\cite{holobrain2026} vision and robot-state encoders. Our method, \textbf{CrossSafe}, is shown in Fig.~\ref{fig:main}.


To enable cross-embodiment safety filtering, we make four key design choices: (i) The actor outputs joint-angle displacements rather than end-effector pose deltas, directly specifying changes to the arms’ joint configurations. (ii) We augment HoloBrain-0's per-link robot tokens 
with safety-related features that were not explicitly encoded by HoloBrain-0, 
including each link’s Cartesian linear and angular velocities and normalized depth in the kinematic tree.
(iii) We let each robot-link token attend to scene patches while explicitly accounting for their spatial relationship in 3D. Particularly, in addition to the standard attention score based on the link and scene features, we bias the score using the distance and direction from the link to each scene patch in a shared world frame. (iv) We design the HJ actor and critic to operate on variable-length sequences of robot link tokens, allowing the same learned modules to be applied across manipulators with different numbers of links, joints, and degrees of freedom.


\subsection{Frozen HoloBrain-0}
Given an observation comprising multi-view RGB-D images, joint angles, gripper openings, and camera intrinsics and extrinsics, the frozen pretrained GroundingDINO variant of HoloBrain-0 produces two token streams: {\em scene tokens} encoding the environment and {\em per-link robot tokens} encoding the robot's configuration. The scene encoder fuses features of the RGB
images generated by a Swin-Transformer  with features of the depth maps generated by another Swin-Transformer, and back-projects
the result through the camera parameters into a shared 3D world frame, yielding
scene tokens $\mathbf{c}=(c_1,\dots,c_{N_{\textit{scene}}})$,
$c_j\in\mathbb R^{256}$ and their corresponding 3D positions $\mathbf{p}^{\textit{scene}}=(p_1^{\textit{scene}},\dots,p_{N_{\textit{scene}}}^{\textit{scene}})$,  $p_j^{\textit{scene}} \in\mathbb R^{3}$.
Forward kinematics converts the joint angles
into link positions and orientations. The robot state encoder embeds these poses together with the states of the  grippers and applies self-attention informed by the robot's kinematic graph, producing the tokens
$\hat{\mathbf{z}}=(\hat z_1,\dots,\hat z_{N_{\textit{link}}})$,
$\hat z_i\in\mathbb R^{256}$. The joint angles are excluded, since 
they are inconsistent across embodiments with different zero-position definitions, rotation directions, and URDFs, while link poses provide a unified geometric reference~\cite{holobrain2026}.
Here $N_{\textit{link}}=N_{\textit{arm}}+N_{\textit{gripper}}$ counts the moving links,
 where $N_{\textit{arm}}$ is the total number of moving arm links, which is equal to the number of actuated arm joints across both arms, 
 and $N_{\textit{gripper}}$ is the number of grippers, each counted as a single link.  
The robot tokens encode the link poses and kinematic structure, while the scene tokens encode the observed environment. 
However, such encoded information might not be sufficient and 
more features might be needed for safety enforcement.

\subsection{Augmenting safety-related features}\label{sec:body-feat}
HoloBrain-0's robot state tokens are extracted from a single observation and do not explicitly encode the link velocities. Velocities are relevant to safety because avoiding contact depends on 
the robot's morphology, its link poses, and 
their motion relative to obstacles. We therefore augment each link token with a safety-related feature vector encoding its velocity, scale-normalized position, orientation, normalized depth in the kinematic tree, and arm identity.

We construct safety-related feature vectors separately for each arm from its URDF and its measured joint angles and velocities, then concatenate the features of the two arms into a single 
sequence for the bimanual robot.  From each URDF, we extract the joint order and axes and estimate each arm's reach $L_{\textit{reach}}$ as the maximum end-effector-to-base distance.
We form one token per link and merge paired gripper fingers into a single token carrying their mean position and the wrist orientation. For link $i$, the safety-related feature vector is 
\begin{equation}
b_i = \Big[\tfrac{p_i^{\textit{link}}-p^{\textit{base}}}{L_{\textit{reach}}},\;
\eta_i^{\textit{wxyz}},\;
\tfrac{J^{\textit{lin}}_i \dot q}{v_0},\;
\tfrac{J^{\textit{ang}}_i \dot q}{\omega_0},\;
\delta_i,\; \chi_i,\; g_i \Big]\in\mathbb R^{16},
\label{eq:body-feat}
\end{equation}
where $p_i^{\textit{link}}$ is the world-frame position of the end of link $i$ that is located at the joint connecting it to its parent, and $p^{\textit{base}}$ is the position of the arm base corresponding to the link, $\eta_i^{\textit{wxyz}}$ is the orientation of link $i$ as a quaternion, $J^{\textit{lin}}_i,J^{\textit{ang}}_i\in\mathbb R^{3\times N_{\textit{arm}}}$ are  the linear and angular velocity Jacobians of link $i$, and $\dot q\in\mathbb R^{N_{\textit{arm}}}$ is the vector of measured joint velocities concatenated across both arms. $\delta_i\in(0,1]$ is the normalized link depth in the kinematic tree and equals 1 for a leaf link, $\chi_i\in\{0,1\}$ is the arm identity for a bimanual manipulator, and $g_i\in[0,1]$ is the normalized gripper opening. 

We use linear and angular link velocities expressed in a common world frame instead of  the joint velocities $\dot q$, because the same $\dot q$ can produce different link motions across arms with different morphologies. 
We express each link position relative to its arm base and divide by $L_{\textit{reach}}$, so that geometrically similar
configurations on arms of different link lengths map to similar values. $v_0=0.5\,\mathrm{m/s}$ and
$\omega_0=\pi\,\mathrm{rad/s}$ are normalization constants. 
The normalized depth $\delta_i$ makes link depth comparable across embodiments.

We fuse each 
token $\hat z_i$ from HoloBrain-0's frozen encoder with our safety-related feature vector $b_i$ from Eq.~\eqref{eq:body-feat} using a fusion module to obtain  $\mathbf{z}^{(0)}=(z_1^{(0)},\dots,z_{N_{\textit{link}}}^{(0)})$, where 
\begin{equation}
z_i^{(0)} = \mathrm{Norm} \!\Big(\mathrm{MLP}\big(
\mathrm{Norm}(\hat z_i)\,\Vert\,b_i\big)\Big)\in\mathbb R^{256},
\label{eq:injection}
\end{equation}
and $\mathrm{MLP}$ is a two-layer network.
The same fusion weights are used for every link across all embodiments.

\subsection{Geometry-aware manipulator--scene attention}
\label{sec:geo-attn}
The link and scene tokens initially encode the robot and the environment separately. To support safety reasoning, each link token needs to encode scene information.
This is important for assessing the spatial relationship between the robot and nearby objects and obstacles.
We connect the two token streams through cross-attention, allowing each link token to encode scene features. 
Standard cross-attention weights scene patches without explicitly accounting for their spatial relation to the link. We instead add a learned attention bias based on the 3D distance and direction from each link to each scene patch, computed from their positions in a shared world frame. This enables the attention mechanism to jointly consider each scene patch’s features and its spatial relationship to each robot link.

For the $\ell$-th cross-attention layer with $H$ heads, 
the queries of the $m$-th head are obtained from the link tokens and the keys and values are obtained from the scene tokens, i.e., 
\begin{equation}
q_i^{(m)} = W_q^{(m)} z_i^{(\ell-1)}, \quad
k_j^{(m)} = W_k^{(m)} c_j, \quad
v_j^{(m)} = W_v^{(m)} c_j,
\label{eq:qkv}
\end{equation}
so each link token is updated to encode scene content while the scene tokens are left unchanged.
For link token $i$ and
scene token $j$, let $r_{ij}=p_j^{\textit{scene}}-p_i^{\textit{link}}$ be the displacement from the link to the patch, $d_{ij}=\lVert r_{ij}\rVert$ be the distance between them, and $\hat r_{ij}=r_{ij}/d_{ij}$ be the normalized displacement representing the direction from the link towards the patch. We bias the attention mechanism with these quantities and augment its output with the direction as follows:
\begin{align}
&\beta^{(m)}_{ij} = \operatorname*{softmax}_j\Big[\tfrac{q_i^{(m)\top}k_j^{(m)}}{\sqrt{C_\textit{head}}}
+ f^{(m)}_\psi(\log d_{ij},\hat r_{ij})\Big], \label{eq:geobias}\\
&o_i = W_o\Big[\ \big\Vert_{m=1}^{H} \textstyle\sum_j \beta^{(m)}_{ij}v^{(m)}_j
\ \Big\Vert\ \big\Vert_{m=1}^{H} \textstyle\sum_j
\beta^{(m)}_{ij}\hat r_{ij}\Big], \label{eq:geoout}
\end{align}
where $C_{\textit{head}}$ is the feature dimension of each attention head, and $f_\psi^{(m)}$ produces the geometric attention bias for head $m$ from the distance and direction between link $i$ and scene patch $j$. The operator $\Vert_{m=1}^{H}$ denotes concatenation across heads, and $W_o$ projects the concatenated features to the same dimension as that of $z_i^{(\ell-1)}$, allowing $o_i$ to be added to $z_i^{(\ell-1)}$ through a residual connection. The second concatenated component in Eq.~\eqref{eq:geoout} is the attention-weighted direction from the link to the scene patches. It explicitly preserves directional information alongside the scene features.

Denoting  Eqs.~\eqref{eq:qkv}--\eqref{eq:geoout} by CA (for cross-attention), 
self-attention over the link tokens by SA, 
a feed-forward layer by FFN,  and $\mathbf{p}^{\textit{link}}=(p_1^{\textit{link}},\dots,p_{N_{\textit{link}}}^{\textit{link}})$, block $\ell=1,2,3$ updates the tokens by:
\begin{equation}
\begin{aligned}
\tilde{\mathbf{z}}^{(\ell)} &= \mathbf{z}^{(\ell-1)} + \mathrm{CA}\big(\mathrm{Norm}(\mathbf{z}^{(\ell-1)}),\mathbf{c}\,,\mathbf{p}^{\textit{link}}\,,\mathbf{p}^{\textit{scene}}\big), \\
\acute{\mathbf{z}}^{(\ell)} &= \tilde{\mathbf{z}}^{(\ell)} + \mathrm{SA}\big(\mathrm{Norm}(\tilde{\mathbf{z}}^{(\ell)})\big), \\
\mathbf{z}^{(\ell)} &= \acute{\mathbf{z}}^{(\ell)} + \mathrm{FFN}\big(\mathrm{Norm}(\acute{\mathbf{z}}^{(\ell)})\big).
\label{eq:block}
\end{aligned}
\end{equation}
The cross-attention aggregates scene-token information into each link token, weighting each scene patch jointly by its feature similarity to the link and by their relative positions.
Self-attention propagates information across link tokens. 
We define the {\em latent state} to be:
\begin{equation}
\mathbf{z} = (z_1^{(3)},\dots,z^{(3)}_{N_{\textit{link}}}).
\end{equation}
Accordingly, we define $\forall  i$, $z_i:=z_i^{(3)}$, for simplicity of notation.



\subsection{Cross-embodiment safety critic and safe policy}
\label{sec:critic-actor}

\paragraph{Safe policy $\pi_{\phi}^{\mathrm{safe}}$}

The safe policy takes the latent state $\mathbf{z}$
as input.
A linear head, shared across all arm joints, maps each joint-associated link token to the mean and log standard deviation of a Gaussian distribution. We sample an action for each joint, squash it to $[-1,1]$ using $\tanh$, and scale it by the corresponding action limit $ a_i^{\max}=\dot q_i^{\lim}\Delta t, $
where $\dot q_i^{\lim}$ is the URDF-provided velocity limit and $\Delta t$ is the control period. 
Gripper tokens do not produce actions as the gripper is controlled only by the nominal policy. 
\paragraph{Hamilton--Jacobi safety critic}
The critic evaluates a candidate action $a=[a_1,\ldots,a_{N_{\textit{arm}}}]^\top$, whose components are joint-angle displacements.
Using the link Jacobians, we compute approximations of corresponding linear and angular displacements as $dp_i = J_i^{\textit{lin}}a \in \mathbb R^3$ and $dw_i = J_i^{\textit{ang}}a \in \mathbb R^3$.

For each arm link $i$, we also denote the command for its connecting joint by $dq_i$. 
To condition the critic on the candidate action, we augment each link token with $dq_i$, $dp_i$, and $dw_i$, which describe how the action would move that link. We concatenate these features with an indicator $\mathbf{1}_i$, which evaluates to one for links whose joints are controlled by the safe policy and to zero otherwise. A  projection layer maps the resulting 8D vector into the token space:
\begin{equation}
\bar z_i^{(0)} = z_i + W_a
[\,dq_i,\,dp_i^\top,\,dw_i^\top,\,\mathbf{1}_i\,]^\top + b_a,
\label{eq:action-inject}
\end{equation}
where $W_a\in\mathbb R^{256\times8}$ and $b_a\in\mathbb R^{256}$ are learned parameters shared across links and embodiments. For gripper tokens, we set $dq_i=\mathbf{1}_i=0$, since the gripper is not controlled by a joint-angle displacement but by a separate opening command issued by the nominal policy; we  retain $dp_i$ and $dw_i$ to describe how the arm moves the gripper.

The critic then applies two additional transformer blocks ($\ell=1,2$) of the form described in Eq.~\eqref{eq:block}. 
Cross-attention combines the action-conditioned tokens with scene information, using the estimated post-action link positions $p_i^{\textit{link}}+dp_i$ to compute the geometric bias in Eq.~\eqref{eq:geobias}. This allows attention to account for how the action would change the distance and direction from each link to each scene patch. Self-attention shares this information across link tokens.

A two-layer feed-forward neural network $g_\theta$, shared across tokens and embodiments, maps each updated token to a safety score $Q_i = g_\theta\big(\bar z_i^{(2)}\big)$, 
where $\bar z_i^{(2)}$ is the updated representation of link token $i$ after the critic's two  transformer blocks. We aggregate scores using a soft minimum: 
\begin{equation}
Q_\theta(\mathbf{z},\mathbf{c},\mathbf{p}^\textit{link},\mathbf{p}^\textit{scene},a)
=
-T\log\sum_i \exp\left(-\frac{Q_i}{T}\right),
\label{eq:softmin}
\end{equation}
where $\mathbf{c}\,,\mathbf{p}^\textit{link}\,,\mathbf{p}^\textit{scene}$ are inputs to the critic’s transformer blocks. 
We write $Q_\theta(\mathbf{z},a)$ from now on for brevity.
Temperature $T>0$ controls the smoothness of the minimum. 
\begin{figure*}[t]
    \centering
    \includegraphics[width=1\linewidth]{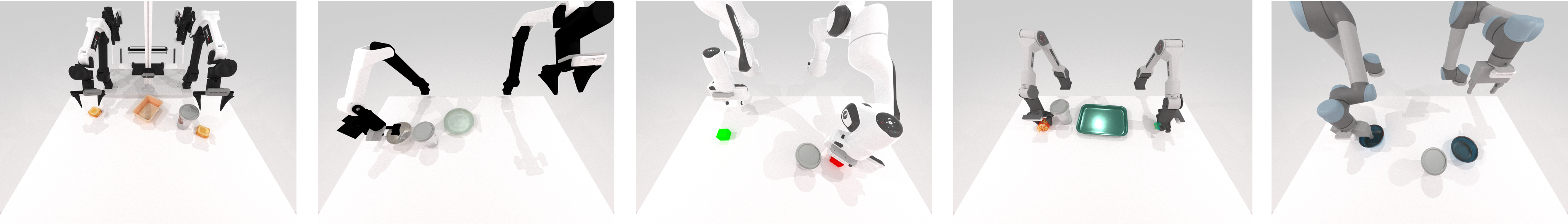}
    \caption{
    Tasks from left to right: Place Bread in Basket, Place Container on Plate, Stack Two Blocks, Place Burger \& Fries, and Stack Two Bowls. Embodiments 
    from left to right: Aloha-AgileX, ARX-X5, Franka-Panda, Piper, and UR5-WSG.}
    \label{fig:robotwin_tasks_robots}
\end{figure*}
\section{Cross-Embodiment Evaluation Setup}



\subsection{Experimental setup} We evaluate CrossSafe in RoboTwin 2.0~\cite{robotwin2} across five bimanual embodiments: Piper, Franka-Panda, ARX-X5, UR5-WSG, and Aloha-AgileX. Franka-Panda has seven DoFs per arm, while the others have six. We use the RoboTwin 2.0 checkpoint of the GroundingDINO variant of HoloBrain-0~\cite{holobrain2026} and keep its encoders frozen. 
We consider the five manipulation tasks shown in Fig.~\ref{fig:robotwin_tasks_robots}, preserving their objectives while adding the same static RoboTwin-OD Box Drink obstacle to each scene. During training, the obstacle is placed along the nominal path with probability $0.7$ and off the path otherwise. During evaluation, it is always placed along the path. The nominal controller is RoboTwin 2.0's cuRobo planner, which plans excluding the added obstacle. 

We define $h(s)$ as the minimum signed distance between the obstacle and the robot and any  object it is grasping. A set of enclosing spheres approximates the robot's body, and a bounding box approximates the obstacle. Thus, $h(s)<0$ indicates overlap between the geometric approximations, which can occur without physical contact.

\subsection{Training pipeline} We jointly optimize all trainable components of CrossSafe shown in
Fig.~\ref{fig:main}. Each model is trained on data from all five tasks and
from its training embodiments, and evaluated on all five embodiments.
A shared warmup buffer contains 20 nominal trajectories per
embodiment-task pair, totaling 500 trajectories, including collision
and collision-free episodes. Each model is trained on the trajectories from
its training embodiments.
All models use the same hyperparameters and collect online trajectories during training at the same  rate.


We use soft actor-critic~\cite{sac} adapted to the HJ reachability case with the loss described in Eq.~(\ref{eq:safety_q_loss}), with a batch size of $64$, a learning rate of $3\times10^{-4}$, and a discount factor of $\gamma=0.9$. We use $T=0.4$. Each model is trained for two days on 
one NVIDIA A40 GPU with 48 GB GPU memory, 8 CPU cores, and 48 GB RAM.  More implementation details and all hyperparameters are in the Appendix.

\subsection{Online safety filtering}

The nominal controller and the safety filter run synchronously at each control step ($25$~Hz). The nominal controller proposes an 
action $a_{\text{nom}}$,  which the safety critic evaluates as
$Q_{\theta}(\mathbf{z},a_{\text{nom}})$.
The executed action is then
\begin{equation}
a =
\begin{cases}
a_{\text{nom}}, & Q_{\theta}(\mathbf{z},a_{\text{nom}})\geq 0,\\
\pi_{\phi}^{\mathrm{safe}}(\mathbf{z}), & Q_{\theta}(\mathbf{z},a_{\text{nom}})<0.
\end{cases}
\end{equation}
Whenever the safety filter intervenes, the nominal controller replans from the state reached after executing the safe action. \yuxuan{On an NVIDIA A40, the HoloBrain-0 vision and robot-state encoders take 
$59.8\pm 1.4$~ms per control step. Our added trainable modules (shown in Fig.~\ref{fig:main}) take $13.1 \pm 0.3$~ms. A full inference for the pipeline in Fig.~\ref{fig:main}  takes $73.0 \pm 1.9$ ms.}

\subsection{Baselines and ablations}

We compare \textbf{CrossSafe} with the unfiltered nominal controller and three ablations: \textbf{CrossSafe w/o Aug} removes the safety-related feature augmentation described in Sec.~\ref{sec:body-feat}. \textbf{CrossSafe w/o Geo} replaces geometry-aware cross-attention with plain cross-attention in the encoder and critic modules. \textbf{CrossSafe w/o Geo, Aug} applies both changes.

We additionally instantiate our method by replacing HoloBrain-0's frozen encoder with that of X-VLA~\cite{xvla2025}, another state-of-the-art VLA. X-VLA encodes observations with a Florence-2 backbone that takes three camera views
and a task instruction. 
Two properties of X-VLA's representation prevent us from directly reusing the fusion module and geometry-aware attention from Sections~\ref{sec:body-feat} and \ref{sec:geo-attn}. First, X-VLA's tokens do not explicitly include metric 3D position information. Second, X-VLA does not provide  the link poses that our safety-related feature vector in Eq.~(\ref{eq:body-feat}) and fusion step in Eq.~(\ref{eq:injection}) are built on. 
It  only exposes the end-effector proprioception. 
In place of the link tokens, the critic and policy operate on two per-arm tokens, obtained by encoding raw end-effector proprioception with a small MLP. These arm tokens attend to the scene tokens through cross-attention, 
structurally analogous to the manipulator-scene attention in Eqs.~(\ref{eq:qkv})-(\ref{eq:block}) 
but without being biased by the geometric information. The actor predicts a bounded change in end-effector position and orientation for each arm, while the critic produces a per-arm safety value pooled by Eq.~(\ref{eq:softmin}), now taken over the two arm tokens rather than over 
link 
tokens. End-effector commands are converted into joint-angle displacements using damped least-squares inverse kinematics and executed when the safety filter intervenes. We call this variant \textbf{X-VLA-Safe}.

\subsection{ Metrics and evaluation protocol}
\label{sec:eval-protocol}
\label{sec:metrics-baselines}

We evaluate each trained model over 50 episodes per task-embodiment pair.  All models are trained across the same five tasks.  In-distribution (InD) evaluation uses embodiments whose data were used to train the model. Out-of-distribution (OOD) evaluation uses embodiments excluded from this training, though they may have appeared during the original training of the  HoloBrain-0 and X-VLA encoders. We use collision rate (CR), success rate (SR), intervention rate (IR), and contact force (Force) as our metrics. CR is the percentage of the 50 evaluation episodes containing at least one physical contact, detected by the simulator, between the manipulator (or an object it holds) and the added obstacle, instead of bounding boxes’ intersections. SR is the percentage of these episodes in which the task is accomplished. To compute IR, we first calculate the percentage of control steps using the safe action within each episode, then average these percentages across the 50 episodes. Force is measured, in newtons, between the manipulator (or an object it holds) and the added obstacle. The simulator reports contact impulses. At each frame, we divide each contact impulse magnitude by the timestep duration to obtain the corresponding force magnitude averaged over that timestep, and take the maximum over these contact points. For each episode, we compute the median of these frame-level values using only frames in which contact occurs, and assign zero to collision-free episodes. Force is then the average of these episode-level values over the 50 episodes.

For each method (X-VLA-Safe; CrossSafe w/o Geo, Aug; CrossSafe w/o Geo; CrossSafe w/o Aug; and CrossSafe), we use five leave-one-embodiment-out folds. Each fold trains a model on four embodiments and evaluates it on those four and the held-out fifth. Since each embodiment is held out in exactly one fold, every task--embodiment pair is evaluated in all five folds: four times as InD and once as OOD.
In Table~\ref{tab:main}, each model is evaluated on 20 InD task--embodiment pairs (1{,}000 episodes) and 5 OOD pairs (250 episodes). \ihabnew{For each model, we average the CR, SR, IR, and Force values computed for each task--embodiment pair across all pairs evaluated by that model, separately for InD and OOD.} For each method, we then report the mean and standard deviation of the averages across its five trained models, \ihabnew{one model per leave-one-embodiment-out fold.} Together, these five models are evaluated over 5{,}000 InD and 1{,}250 OOD episodes.

Table~\ref{tab:specialist} compares the five CrossSafe models from Table~\ref{tab:main}, termed \emph{generalists}, with five CrossSafe \emph{specialists}. Each generalist is trained on a different combination of four embodiments, whereas each specialist is trained on one embodiment. Each specialist is evaluated on its training embodiment (InD) and the other four embodiments (OOD), giving 5 InD task--embodiment pairs (250 episodes) and 20 OOD pairs (1{,}000 episodes). Each generalist is evaluated on its four training embodiments (InD) and the held-out fifth embodiment (OOD), giving 20 InD pairs (1{,}000 episodes) and 5 OOD pairs (250 episodes). For each model, we average the metrics across its evaluated task--embodiment pairs, separately for InD and OOD. 
\ihabnew{We then report the mean and standard deviation of these averages separately across the five generalists and the five specialists.}
Specialists are evaluated over 1{,}250 InD and 5{,}000 OOD episodes, and generalists over 5{,}000 InD and 1{,}250 OOD episodes.

Detailed results for each method on every task--embodiment pair, separately for InD and OOD evaluations, are in the Appendix.

\section{Results}

\begin{table*}[t]
\centering
\tiny
\setlength{\tabcolsep}{5pt}
\renewcommand{\arraystretch}{1.15}
\begin{tabular}{l cccc cccc}
\toprule
& \multicolumn{4}{c}{In-distribution Embodiments} & \multicolumn{4}{c}{Out-of-distribution Embodiment} \\
\cmidrule(lr){2-5}\cmidrule(lr){6-9}
Method & CR (\%)\,$\downarrow$ & SR (\%)\,$\uparrow$ & IR (\%) & Force [N]\,$\downarrow$
       & CR (\%)\,$\downarrow$ & SR (\%)\,$\uparrow$ & IR (\%) & Force [N] \,$\downarrow$ \\
\midrule
\textit{Nominal} & 64.1 & 32.2 & --- & 179.6
                 & 64.1 & 32.2 & --- & 179.6 \\
X-VLA-Safe & 50.4\,\tiny{$\pm$0.7} & 17.0\,\tiny{$\pm$4.1} & 5.2\,\tiny{$\pm$0.8} & 189.7\,\tiny{$\pm$30.3}
      & 53.8\,\tiny{$\pm$17.4} & 16.7\,\tiny{$\pm$12.1} & 4.8\,\tiny{$\pm$1.5} & 247.0\,\tiny{$\pm$144.6} \\
\midrule
CrossSafe w/o Geo, Aug & 38.3\,\tiny{$\pm$7.7} & 31.3\,\tiny{$\pm$4.1} & 11.1\,\tiny{$\pm$7.0} & 81.3\,\tiny{$\pm$18.0}
                   & 51.5\,\tiny{$\pm$18.6} & 35.0\,\tiny{$\pm$15.7} & 7.6\,\tiny{$\pm$4.3} & 126.2\,\tiny{$\pm$94.8} \\
CrossSafe w/o Geo      & 45.4\,\tiny{$\pm$7.4} & 23.8\,\tiny{$\pm$5.9} & 20.2\,\tiny{$\pm$4.6} & 60.2\,\tiny{$\pm$17.4}
                   & 53.4\,\tiny{$\pm$17.7} & 30.0\,\tiny{$\pm$11.5} & 15.5\,\tiny{$\pm$10.4} & 79.0\,\tiny{$\pm$53.7} \\
CrossSafe w/o Aug      & 37.1\,\tiny{$\pm$5.0} & 26.0\,\tiny{$\pm$7.5} & 7.7\,\tiny{$\pm$1.8} & 150.1\,\tiny{$\pm$35.8}
                   & 49.7\,\tiny{$\pm$13.3} & 23.3\,\tiny{$\pm$13.1} & 9.2\,\tiny{$\pm$8.8} & 133.5\,\tiny{$\pm$31.8} \\
CrossSafe    & 39.8\,\tiny{$\pm$3.8} & 23.7\,\tiny{$\pm$2.8} & 13.2\,\tiny{$\pm$1.2} & 113.6\,\tiny{$\pm$41.9}
                   & 49.8\,\tiny{$\pm$8.3} & 29.4\,\tiny{$\pm$15.7} & 9.2\,\tiny{$\pm$6.1} & 102.9\,\tiny{$\pm$44.3} \\
\bottomrule
\end{tabular}
\caption{Comparison of methods using CR, SR, IR, and Force. Each learned method has five trained models, one per leave-one-embodiment-out fold. Each model is evaluated on its four training embodiments (InD) and held-out fifth (OOD).}
\label{tab:main}
\end{table*}

\begin{table}[t]
\centering
\tiny
\setlength{\tabcolsep}{4pt}
\renewcommand{\arraystretch}{1.15}
\begin{tabular}{ll cccc}
\toprule
& Model & CR (\%)\,$\downarrow$ & SR (\%)\,$\uparrow$ & IR (\%) & Force [N]\,$\downarrow$ \\
\midrule
\multirow{2}{*}{InD Embodiment(s)}
  & Specialist & \textbf{37.6}\,\tiny{$\pm$21.3} & 18.7\,\tiny{$\pm$10.2} & 11.2\,\tiny{$\pm$4.8} & \textbf{113.0}\,\tiny{$\pm$87.1} \\
  & Generalist & 39.8\,\tiny{$\pm$3.8}  & \textbf{23.7}\,\tiny{$\pm$2.8}  & 13.2\,\tiny{$\pm$1.2} & 113.6\,\tiny{$\pm$41.9} \\
\midrule
\multirow{2}{*}{OOD Embodiment(s)}
  & Specialist & 54.4\,\tiny{$\pm$8.2}  & 27.0\,\tiny{$\pm$5.5}  & 5.6\,\tiny{$\pm$3.4} & 144.8\,\tiny{$\pm$24.5} \\
  & Generalist & \textbf{49.8}\,\tiny{$\pm$8.3}  & \textbf{29.4}\,\tiny{$\pm$15.7} & 9.2\,\tiny{$\pm$6.1} & \textbf{102.9}\,\tiny{$\pm$44.3} \\
\bottomrule
\end{tabular}
\caption{CrossSafe specialists vs. generalists.}
\label{tab:specialist}
\end{table}




\paragraph{A single filter improves safety across multiple tasks and embodiments}

From Table~\ref{tab:main}, CrossSafe reduces the collision rate (CR) of the unfiltered nominal planner from
$64.1\%$ to $39.8\%$ on in-distribution embodiments (InD) and to $49.8\%$ on the out-of-distribution
embodiment (OOD). The Force metric also falls from 179.6 N to 113.6 N InD and 102.9 N OOD.  Across the 25 task--embodiment pairs, CrossSafe lowers CR
relative to the nominal policy in $21/25$ InD and in $19/25$ OOD pairs. 
Relative to the nominal controller, CrossSafe reduces the mean CR over the five tasks for every in-distribution embodiment. CR decreases on all five tasks for ARX-X5 (mean CR decreases from $75.2\%$ to $42.5\%$) and Aloha-AgileX ($66.4\%$ to $13.3\%$), on 4 of 5 tasks for Piper ($51.6\%$ to $29.3\%$) and Franka-Panda ($66.0\%$ to $57.1\%$), and on 3 of 5 tasks for UR5-WSG ($61.2\%$ to $56.9\%$). For OOD evaluations, mean CR decreases on four of the five held-out
embodiments. CR decreases on 5 of 5 tasks for ARX-X5 (mean CR decreases from
$75.2\%$ to $40.0\%$) and Aloha-AgileX ($66.4\%$ to $44.4\%$), and on 4
of 5 for Franka-Panda ($66.0\%$ to $60.4\%$) and UR5-WSG ($61.2\%$ to $48.4\%$), while on Piper mean CR increases from $51.6\%$ to
$56.0\%$, with CR reduced on only 1 of 5 \ihabnew{tasks.}


\paragraph{CrossSafe generalizes to embodiments unseen during safety-filter training}
On the OOD embodiment, CrossSafe reduces CR by 14.3 percentage points and Force from $179.6\,$N to $102.9\,$N
(Table~\ref{tab:main}), while intervening on only $9.2\%$ of control steps.
Among the held-out folds, Franka-Panda provides the clearest test of generalization across embodiments.
 Franka-Panda has seven DoFs per arm
while the other four embodiments have six, so in that fold every trainable
component of the filter is trained exclusively on 6-DoF arms and the safe policy must then output an additional joint angle displacement per arm at test time. Despite these differences in robot morphology and action dimension, CrossSafe still lowers mean CR from $66.0\%$ to $60.4\%$ (it lowers it on $4$ out of $5$ \ihabnew{tasks}) 
while raising \ihabnew{mean} SR from $40.4\%$ to
$45.2\%$, and attains the lowest CR of the five \ihabnew{methods} on the held-out Franka-Panda embodiment. 
When the Franka-Panda embodiment is held out, CrossSafe's three ablation variants (CrossSafe w/o Geo, Aug; CrossSafe w/o Geo; CrossSafe w/o Aug) all raise mean CR above the nominal controller’s CR (from $66.0\%$ to $76.0\%$, $79.2\%$, and $66.4\%$, respectively). These results demonstrate that the different components of CrossSafe helped it  generalize from training embodiments with six DoFs per arm to an unseen embodiment with seven. 

\paragraph{Geometry-aware attention and safety-feature augmentation enable  generalization} Table~\ref{tab:main} compares CrossSafe with CrossSafe w/o Geo, CrossSafe w/o Aug, and CrossSafe w/o Geo, Aug to assess geometry-aware attention and safety-feature augmentation. All four CrossSafe variants achieve lower mean CR and Force than the nominal controller in both InD and OOD evaluations.  When safety-feature augmentation is omitted, CrossSafe w/o Aug achieves lower mean OOD CR than CrossSafe w/o Geo, Aug. When it is included, CrossSafe achieves lower mean OOD CR than CrossSafe w/o Geo. In both comparisons, the model using geometry-aware attention achieves lower mean OOD CR than its counterpart using standard cross-attention.
With geometry-aware attention retained in both models, CrossSafe w/o Aug and CrossSafe achieve nearly identical mean OOD CRs ($49.7\%$ and $49.8\%$, respectively), while CrossSafe achieves a higher SR ($29.4\%$ vs.\ $23.3\%$), lower Force ($102.9\,$N vs.\ $133.5\,$N), and the same mean IR.
These results support our hypothesis that combining geometry-aware attention and safety-feature augmentation benefits generalization to unseen embodiments.


\paragraph{CrossSafe achieves lower collision rates and higher task success rates than X-VLA-Safe}
Compared with X-VLA-Safe, CrossSafe achieves lower mean CR and Force and higher mean SR in both
InD and OOD evaluations (Table~\ref{tab:main}).
Relative to nominal, CrossSafe cuts CR by $37.9\%$ and $22.3\%$ and contact force by $36.7\%$ and
$42.7\%$ for InD and OOD, respectively, while giving up only $26.4\%$ and $8.7\%$ of task success.
X-VLA-Safe reduces CR by only $21.4\%$ InD and $16.1\%$ OOD, and does so while halving task success
($47.2\%$ and $48.1\%$ drops, respectively) and \emph{raising} Force $5.6\%$ and $37.5\%$
above nominal.




\paragraph{Training across more embodiments improves generalization to OOD settings with minimal effects on InD performance}
Compared to specialists, the generalists improve all OOD outcome metrics: CR $49.8\%$ vs.\ $54.4\%$, SR $29.4\%$ vs.\ $27.0\%$,
and Force $102.9\,$N vs.\ $144.8\,$N. Jointly training the safety filter across embodiments therefore yields a
filter that exhibits better generalization. 
For the InD task--embodiment pairs, against the specialists, the generalists attain lower CR in 13 of 25  and higher SR in 15 of 25. The generalists' mean InD SR is also higher ($23.7\%$ vs.\ $18.7\%$), although mean InD CR is slightly higher ($39.8\%$ vs.\ $37.6\%$).



\section{Conclusion}

We hypothesized that the reasoning required to satisfy a safety
constraint is largely shared across robots, while the action that realizes
it depends on each robot's morphology, kinematics, and dynamics. CrossSafe instantiates this idea with a representation of the robot as a
variable-length sequence of link tokens encoding kinematics and nearby
scene geometry, allowing one HJ critic and safe policy to be learned
across bimanual manipulators with different degrees of freedom. 
The representation itself is not tied to the HJ formulation and can be used to design  
other latent safety filters, including ones based on neural control barrier functions. 
In RoboTwin~2.0, CrossSafe reduces the nominal policy's
collision rate and contact force both on training embodiments and on
embodiments unseen during the training of the safety filter.
Two limitations remain. As a learned filter, CrossSafe provides no formal
guarantee, and filtered collision rates remain relatively high. Future work includes
calibrating the learned value function using conformal prediction or
scenario optimization to obtain probabilistic guarantees, training across a larger and more diverse set of 
robot embodiments and tasks, and hardware validation.

\section{Acknowledgements}
This research used both the DeltaAI advanced computing and data resource, which is supported by the National Science Foundation (award OAC 2320345) and the State of Illinois, and the Delta advanced computing and data resource which is supported by the National Science Foundation (award OAC 2005572) and the State of Illinois. Delta and DeltaAI are joint efforts of the University of Illinois Urbana-Champaign and its National Center for Supercomputing Applications. Access to these resources was provided through an allocation from the NSF ACCESS program.

\bibliographystyle{IEEEtran}
\bibliography{references}

%
%
%
 
\clearpage
\section*{Appendix}

\setcounter{subsection}{0}
\setcounter{subsubsection}{0}

\renewcommand{\thesubsection}{\arabic{subsection}}
\renewcommand{\thesubsubsection}{\thesubsection.\arabic{subsubsection}}

\etocsettocstyle
  {}
  {\par\vspace{6pt}}

\etocsetstyle{subsection}
  {}
  {}
  {\noindent\textbf{\etocnumber\quad\etocname}
   \dotfill\etocpage\par}
  {}

\etocsetstyle{subsubsection}
  {}
  {}
  {\noindent\hspace{1em}\etocnumber\quad\etocname
   \dotfill\etocpage\par}
  {}

\localtableofcontents

\subsection{Simulation environment}
\label{app:env}
\subsubsection{Control frequency}

All experiments run in RoboTwin~2.0 on top of SAPIEN. The physics timestep
is fixed at $250$~Hz and the control loop at $25$~Hz, so one control tick is
exactly ten physics substeps and the control period is $\Delta t = 0.04$~s. At each control tick, the critic receives the nominal action $a_\textit{nom}$, defined as the displacement from the current measured joint configuration to the one the nominal plan reaches 0.04~s later, or to its final configuration when less than one tick remains.
Episodes are capped at $450$ control ticks ($18$~s of simulated time).

\subsubsection{Embodiments}

All embodiments are
dual-arm configurations of the corresponding RoboTwin~2.0 robot, with the
two arms treated as one system. Franka-Panda has seven actuated joints per
arm. The remaining four have six.
Joint action limits are derived from the URDF of each robot:
$a^{\max}_i = \dot q^{\lim}_i \Delta t $.


\subsubsection{Tasks}

We use five bimanual RoboTwin~2.0 manipulation tasks: \emph{Place Bread in
Basket}, \emph{Place Container on Plate}, \emph{Stack Two Blocks},
\emph{Place Burger \& Fries}, and \emph{Stack Two Bowls}. Task objectives,
success criteria, and scene randomization are inherited unchanged from RoboTwin~2.0. Background randomization is left at the benchmark defaults and turned off. Our only modification is the addition of one static obstacle per scene, described next.

\subsubsection{Obstacle placement}

The obstacle is the RoboTwin-OD \texttt{068\_boxdrink} mesh, a box of roughly $11.0 \times 15.4 \times 11.6$~cm, spawned as a static actor resting on the table. Let $p_{\text{pick}}$ and $p_{\text{place}}$ be the table-plane positions of the picked object and the place target. The obstacle is centered at
\begin{equation}
p(\rho) = (1-\rho)\, p_{\text{pick}} + \rho\, p_{\text{place}}, \qquad
\rho \sim \mathcal{U}[0.22, 0.48],
\end{equation}
drawn once per episode, so that $\rho = 0$ is the pick pose and $\rho = 1$ the place pose.

In \emph{on-path} mode the obstacle sits at $p(\rho)$; in \emph{off-path} mode
it is offset perpendicular to the segment. During collection the mode is drawn
independently of the task and embodiment, \emph{on-path} with probability $0.7$, so
that filter engagement is not confounded with the obstacle being in the way.
All evaluation scenes are \emph{on-path}.

\subsubsection{Failure function}

The failure function is the signed distance between the robot system and
the obstacle,
\begin{equation}
h(s) = \min \; \mathrm{dist}\!\left(\mathcal{B}(s) \cup \mathcal{P}(s),\;
\mathcal{O}\right),
\end{equation}
where $\mathcal{B}(s)$ is the set of collision spheres on the
moving links of both arms, $\mathcal{P}(s)$ is the grasped object if one is
held, and $\mathcal{O}$ is the oriented bounding box of the obstacle.

\subsection{Safety critic and policy: implementation details}
\label{app:heads}

\subsubsection{Soft actor-critic training}

We instantiate two critic heads (twin critics) \cite{twincrit} with separate  transformer blocks and separate value heads, both using the same encoded latent state.  We train the critics and the safe policy with SAC~\cite{haarnoja2018soft} adapted to the HJ reachability setting: the critics regress the discounted avoid target $y_t$ of Eq.~\eqref{eq:safety_q_loss}. The entropy loss is added to the actor loss, where it acts as an exploration regularizer with $\alpha$ annealed from $0.2$ to $0.02$ over the first $20{,}000$ gradient steps.  Target networks are Polyak averaged with $\tau = 0.005$.

\subsubsection{Safe actor head}

The actor is a single linear map $\mathrm{Linear}(256 \to 2)$ applied to
every actuated link token, producing a mean and a log standard deviation.
The log standard deviation is clamped to $[-5, 0]$. Actions are sampled with
the reparameterization trick, squashed by $\tanh$, scattered into the action
vector by joint index, and only then scaled by the per-joint limit
$a^{\max}_i$, so $|a_i| \le a^{\max}_i$ holds by construction.


\subsubsection{Data collection}

We first collect a single shared warmup buffer using only the nominal controller, cycling through a fixed, shuffled ordering of the $25$ task–embodiment pairs. Collection continues until $20$ successful episodes are retained for each pair, yielding $500$ trajectories in total.

The same warmup buffer is reused across all training splits and ablations, with each model sampling only trajectories from its training embodiments.

Training then alternates between $1024$ gradient steps and the collection of three fresh episodes with the current weights. Within a collection round, an episode runs with the safety filter active with probability $0.8$
and nominal-only otherwise.

\subsubsection{Parameter count}

Table~\ref{tab:params} reports trainable parameter counts. The frozen HoloBrain-0 encoders are excluded; they receive no gradient.

\begin{table}[t]
\centering
\caption{Trainable parameters (frozen HoloBrain-0 excluded).}
\label{tab:params}
\small
\begin{tabular}{lr}
\toprule
Module & Parameters \\
\midrule
Fusion of per-link tokens & 202{,}752 \\
Geometry-aware transformer ($3$ blocks)                       & 3{,}177{,}720 \\
Safe actor head                                         & 514 \\
Twin safety critic ($2 \times$ [$2$ blocks + value head])& 4{,}373{,}666 \\
\midrule
\textbf{CrossSafe (total)}                              & \textbf{7{,}754{,}652} \\
\midrule
CrossSafe w/o Geo                                       & 7{,}708{,}676 \\
CrossSafe w/o Aug                                       & 7{,}750{,}556 \\
CrossSafe w/o Geo, Aug                                  & 7{,}704{,}580 \\
\bottomrule
\end{tabular}
\end{table}

\subsubsection{CrossSafe training algorithm}
Algorithm \eqref{alg:train} shows how the safe actor and critic are trained.

\begin{algorithm}[t]
\caption{CrossSafe training}
\label{alg:train}
\begin{algorithmic}[1]
\REQUIRE embodiment pool $\mathcal{E}$, shared warmup buffer $\mathcal{D}_0$
\STATE $\mathcal{D} \gets$ transitions of $\mathcal{D}_0$ whose embodiment is in $\mathcal{E}$
\STATE initialize encoder $\psi$, critic $\theta$, actor $\phi$
\STATE target copies $\psi^- \gets \psi$, \; $\theta^- \gets \theta$
\FOR{round $r = 1$ \TO $R$}
  \FOR{$1024$ gradient steps}
    \STATE sample a batch of $64$ transitions from $\mathcal{D}$
    \STATE encode $z \gets E_\psi(s)$ and $z' \gets E_{\psi^-}(s')$
    \STATE update $\theta$ and $\psi$ using Eq. \eqref{eq:safety_q_loss}
    \STATE update $\phi$ to maximize $Q_\theta(z, \pi_\phi(z))$ and actor
    entropy, holding $\theta$ and $\psi$ fixed
    \STATE $\psi^- \gets (1-\tau)\psi^- + \tau\psi$; \;
           $\theta^- \gets (1-\tau)\theta^- + \tau\theta$
  \ENDFOR
  \STATE save checkpoint
  \FOR{$3$ episodes, cycling over tasks and embodiments in $\mathcal{E}$}
    \STATE with probability $0.8$ roll out with the safety filter, else roll
           out the nominal controller with random action perturbations
    \STATE append the episode's transitions to $\mathcal{D}$
  \ENDFOR
\ENDFOR
\end{algorithmic}
\end{algorithm}
\subsubsection{Hyperparameters}

Table~\ref{tab:hyper} lists every hyperparameter. All methods, ablations, folds, and specialists use identical values; the only differences across runs are the embodiment pool used for training the safety filter components.

\begin{table}[ht]
\centering
\caption{Hyperparameters.}
\label{tab:hyper}
\scriptsize
\begin{tabular}{lc}
\toprule
\multicolumn{2}{l}{\emph{Problem definition}} \\
\midrule
Physics frequency                        & $250$~Hz \\
Control frequency                        & $25$~Hz \\
Control period $\Delta t$                & $0.04$~s \\
Action-limit fraction $\kappa$           & $1.0$ \\
Soft-min temperature                     & $T = 0.4$  \\
Discount $\gamma$                        & $0.9$  \\
Max episode length                       & $450$ control steps ($18$~s) \\
Obstacle model                           & \texttt{068\_boxdrink} \\
\emph{Off-path} fraction (data collection)           & $0.3$ \\
\midrule
\multicolumn{2}{l}{\emph{Architecture}} \\
\midrule
link token width                              & $256$ \\
Attention heads                          & $8$ \\
Feed-forward width                       & $1024$ \\
Normalization                            & RMSNorm  \\
Activation                               & SiLU \\
Scene tokens $N_{\text{scene}}$          & $1200$ ($3$ cams $\times\ 400$) \\
Image resolution                         & $320 \times 256$ \\
\midrule
\multicolumn{2}{l}{\emph{Optimization}} \\
\midrule
Optimizer                                & Adam \\
Learning rate                            & $3 \times 10^{-4}$ \\
Batch size                               & $64$ \\
Target Polyak $\tau$                     & $0.005$ \\
Entropy coefficient $\alpha$             & $0.2 \to 0.02$ over $20$k steps \\
Gradient-norm clip                       & $10$ \\
Gradient steps per round                 & $1024$ \\
Episodes collected per round             & $3$ \\
Action perturbation probability                 & $0.05$ \\
Warmup trajectories                      & $500$ ($20$ per task-embodiment pair) \\
\bottomrule
\end{tabular}
\end{table}


\subsection{Evaluation}
\label{app:eval}

\subsubsection{Scene determinism}

All methods, all ablations, and the unfiltered nominal controller are evaluated on identical scenes, and re-running an evaluation reproduces the same results. Each (task, embodiment) cell logs exactly $50$ episodes: planner failures that produce an empty trace, and obstacle-spawn failures, redraw a fresh deterministic scene from a derived seed until a real episode completes, with a cap on attempts. The safe policy is evaluated deterministically, using the pre-activation mean with no sampling.

\subsubsection{Extended results}
\label{app:extended}

Tables~\ref{tab:app1-ind} and~\ref{tab:app2-ood} report collision and success rate for every task--embodiment pair.

\begin{table*}[ht]
\centering
\small
\caption{In-distribution collision and success rate per task and embodiment,
shown as CR/SR (both \%). \emph{Nominal} is the unfiltered cuRobo planner and
is training-independent. \emph{X-VLA-Safe} and the four \emph{CrossSafe}
variants are the leave-one-embodiment-out models: each cell averages the four
models that had that embodiment in their training pool ($4 \times 50 = 200$
episodes). \emph{Specialist} is a CrossSafe model trained on that embodiment
alone and evaluated on it ($50$ episodes). Among the learned methods, bold
marks the lowest CR and the highest SR in each row. \emph{Task avg.} averages
the five embodiments; \emph{Overall average} averages all $25$ pairs.}
\label{tab:app1-ind}
\begin{tabular}{ll ccccccc}
\toprule
Task & Emb. & \textit{Nominal} & X-VLA & Specialist & \shortstack{CrossSafe\\w/o Geo, Aug} & \shortstack{CrossSafe\\w/o Geo} & \shortstack{CrossSafe\\w/o Aug} & \textbf{CrossSafe} \\
\midrule
Place Bread in Basket & Piper & 74.0/28.0 & 46.0/8.5 & 38.0/16.0 & 47.5/15.0 & 53.5/\textbf{17.0} & 48.5/16.5 & \textbf{37.5}/15.0 \\
 & Franka-Panda & 98.0/10.0 & 85.5/4.0 & 66.0/0.0 & 62.5/\textbf{7.0} & 74.0/4.5 & \textbf{60.0}/1.5 & 80.0/0.5 \\
 & ARX-X5 & 88.0/24.0 & 66.0/5.5 & \textbf{30.0}/6.0 & 40.0/\textbf{34.5} & 63.0/26.0 & 46.5/25.0 & 63.5/14.5 \\
 & UR5-WSG & 98.0/10.0 & 87.0/12.0 & 96.0/10.0 & \textbf{77.0}/\textbf{46.5} & 82.0/34.0 & 83.5/27.5 & 93.0/16.5 \\
 & Aloha-AgileX & 76.0/16.0 & 73.0/6.0 & \textbf{0.0}/0.0 & 43.0/\textbf{12.0} & 51.0/7.0 & 44.5/5.0 & 13.0/0.0 \\
 & \textit{Task avg.} & 86.8/17.6 & 71.5/7.2 & \textbf{46.0}/6.4 & 54.0/\textbf{23.0} & 64.7/17.7 & 56.6/15.1 & 57.4/9.3 \\
\cmidrule(lr){1-9}
Place Container on Plate & Piper & 38.0/0.0 & 12.0/1.5 & \textbf{10.0}/\textbf{18.0} & 15.5/11.5 & 32.5/6.0 & \textbf{10.0}/13.5 & 25.5/7.0 \\
 & Franka-Panda & 86.0/8.0 & 51.5/\textbf{28.0} & 48.0/20.0 & 53.5/25.0 & 62.5/21.5 & \textbf{46.5}/19.5 & 69.0/25.0 \\
 & ARX-X5 & 76.0/2.0 & 45.0/9.5 & 28.0/18.0 & \textbf{18.5}/\textbf{37.0} & 42.5/20.5 & 33.5/21.0 & 42.0/14.0 \\
 & UR5-WSG & 70.0/4.0 & 47.0/3.0 & 50.0/16.0 & 44.5/47.5 & 37.0/\textbf{48.5} & \textbf{35.0}/26.5 & 48.5/25.5 \\
 & Aloha-AgileX & 76.0/6.0 & 42.0/12.0 & 28.0/10.0 & \textbf{17.0}/\textbf{36.5} & 33.0/13.0 & 23.5/16.5 & 26.5/11.0 \\
 & \textit{Task avg.} & 69.2/4.0 & 39.5/10.8 & 32.8/16.4 & 29.8/\textbf{31.5} & 41.5/21.9 & \textbf{29.7}/19.4 & 42.3/16.5 \\
\cmidrule(lr){1-9}
Stack Two Blocks & Piper & 54.0/60.0 & 57.0/32.5 & \textbf{20.0}/\textbf{62.0} & 30.5/59.0 & 38.5/43.5 & \textbf{20.0}/57.5 & 41.0/53.5 \\
 & Franka-Panda & 32.0/90.0 & 45.5/60.0 & 42.0/46.0 & 53.0/61.0 & 42.5/53.0 & \textbf{26.5}/\textbf{64.0} & 40.5/58.5 \\
 & ARX-X5 & 82.0/72.0 & 66.5/40.0 & 60.0/38.0 & 48.0/\textbf{65.5} & 55.5/63.0 & 56.5/50.5 & \textbf{47.0}/61.0 \\
 & UR5-WSG & 24.0/84.0 & 31.5/72.5 & 50.0/62.0 & 35.5/81.0 & 35.5/78.5 & 38.5/\textbf{83.5} & \textbf{31.0}/80.5 \\
 & Aloha-AgileX & 80.0/44.0 & 64.0/1.5 & 28.0/0.0 & 29.5/\textbf{2.0} & 40.5/0.0 & 29.5/1.0 & \textbf{11.5}/0.0 \\
 & \textit{Task avg.} & 54.4/70.0 & 52.9/41.3 & 40.0/41.6 & 39.3/\textbf{53.7} & 42.5/47.6 & \textbf{34.2}/51.3 & \textbf{34.2}/50.7 \\
\cmidrule(lr){1-9}
Place Burger \& Fries & Piper & 84.0/18.0 & 38.0/14.5 & \textbf{24.0}/40.0 & 30.5/36.5 & 51.5/27.5 & 27.0/\textbf{46.0} & 31.0/31.0 \\
 & Franka-Panda & 100.0/6.0 & 79.0/0.5 & \textbf{66.0}/0.0 & 83.5/0.5 & 74.0/1.0 & 70.0/0.0 & 83.0/\textbf{1.5} \\
 & ARX-X5 & 100.0/0.0 & 61.0/1.5 & 58.0/0.0 & \textbf{32.5}/\textbf{16.5} & 47.5/8.5 & 42.0/5.5 & 51.0/2.5 \\
 & UR5-WSG & 100.0/2.0 & 88.0/8.5 & 100.0/4.0 & 88.5/\textbf{30.0} & 93.5/17.0 & \textbf{86.0}/25.0 & 95.0/7.0 \\
 & Aloha-AgileX & 70.0/6.0 & 64.0/0.0 & 20.0/0.0 & 25.5/\textbf{7.0} & 31.0/0.0 & 39.0/3.0 & \textbf{15.5}/0.0 \\
 & \textit{Task avg.} & 90.8/6.4 & 66.0/5.0 & 53.6/8.8 & \textbf{52.1}/\textbf{18.1} & 59.5/10.8 & 52.8/15.9 & 55.1/8.4 \\
\cmidrule(lr){1-9}
Stack Two Bowls & Piper & 8.0/28.0 & 12.5/2.0 & 10.0/\textbf{10.0} & 7.0/9.0 & 14.5/4.0 & \textbf{5.0}/8.0 & 11.5/6.5 \\
 & Franka-Panda & 14.0/88.0 & \textbf{7.5}/\textbf{55.0} & 12.0/42.0 & 24.5/40.5 & 20.0/35.0 & 17.5/51.0 & 13.0/54.5 \\
 & ARX-X5 & 30.0/66.0 & 44.0/23.0 & \textbf{4.0}/26.0 & 16.5/46.5 & 20.0/31.0 & 16.0/36.5 & 9.0/\textbf{47.5} \\
 & UR5-WSG & 14.0/86.0 & 21.5/22.5 & 48.0/24.0 & 23.5/56.0 & 27.5/35.5 & \textbf{17.0}/44.5 & \textbf{17.0}/\textbf{58.5} \\
 & Aloha-AgileX & 30.0/48.0 & 24.0/\textbf{1.0} & 4.0/0.0 & 10.0/0.0 & 11.5/0.5 & 4.5/0.5 & \textbf{0.0}/0.0 \\
 & \textit{Task avg.} & 19.2/63.2 & 21.9/20.7 & 15.6/20.4 & 16.3/30.4 & 18.7/21.2 & 12.0/28.1 & \textbf{10.1}/\textbf{33.4} \\
\midrule
\multicolumn{2}{l}{\textbf{Overall average}} & 64.1/32.2 & 50.4/17.0 & 37.6/18.7 & 38.3/\textbf{31.3} & 45.4/23.8 & \textbf{37.1}/26.0 & 39.8/23.7 \\
\bottomrule
\end{tabular}
\end{table*}

\begin{table*}[ht]
\centering
\small
\caption{Zero-shot collision and success rate on held-out embodiments, shown
as CR/SR (both \%). Conditions are as in Table~\ref{tab:app1-ind}, but each
\emph{X-VLA-Safe} and \emph{CrossSafe} cell comes from the single model for
which that embodiment was excluded from safety-filter training ($50$
episodes), and each \emph{Specialist} cell averages the four
single-embodiment models that did not train on it ($4 \times 50 = 200$
episodes). Among the learned methods, bold marks the lowest CR and the
highest SR in each row.}
\label{tab:app2-ood}
\begin{tabular}{ll ccccccc}
\toprule
Task & Emb. & \textit{Nominal} & X-VLA & Specialist & \shortstack{CrossSafe\\w/o Geo, Aug} & \shortstack{CrossSafe\\w/o Geo} & \shortstack{CrossSafe\\w/o Aug} & \textbf{CrossSafe} \\
\midrule
Place Bread in Basket & Piper & 74.0/28.0 & \textbf{24.0}/6.0 & 37.0/9.5 & 68.0/12.0 & 76.0/\textbf{26.0} & 64.0/16.0 & 74.0/18.0 \\
 & Franka-Panda & 98.0/10.0 & \textbf{94.0}/18.0 & 95.5/11.5 & 98.0/\textbf{22.0} & 98.0/6.0 & 98.0/10.0 & 96.0/6.0 \\
 & ARX-X5 & 88.0/24.0 & 64.0/4.0 & 66.5/25.0 & \textbf{32.0}/40.0 & 36.0/28.0 & 38.0/18.0 & 42.0/\textbf{44.0} \\
 & UR5-WSG & 98.0/10.0 & 82.0/8.0 & 82.0/9.0 & 92.0/4.0 & \textbf{66.0}/\textbf{22.0} & 72.0/4.0 & 82.0/0.0 \\
 & Aloha-AgileX & 76.0/16.0 & 82.0/8.0 & \textbf{24.5}/0.0 & 60.0/8.0 & 54.0/\textbf{22.0} & 60.0/4.0 & 66.0/4.0 \\
 & \textit{Task avg.} & 86.8/17.6 & 69.2/8.8 & \textbf{61.1}/11.0 & 70.0/17.2 & 66.0/\textbf{20.8} & 66.4/10.4 & 72.0/14.4 \\
\cmidrule(lr){1-9}
Place Container on Plate & Piper & 38.0/0.0 & \textbf{18.0}/0.0 & 47.0/2.5 & 34.0/2.0 & 42.0/2.0 & 48.0/2.0 & 52.0/\textbf{4.0} \\
 & Franka-Panda & 86.0/8.0 & 76.0/20.0 & 91.5/21.0 & 84.0/50.0 & 98.0/20.0 & 78.0/20.0 & \textbf{70.0}/\textbf{58.0} \\
 & ARX-X5 & 76.0/2.0 & 54.0/8.0 & 58.5/20.5 & 30.0/\textbf{44.0} & 38.0/38.0 & 34.0/12.0 & \textbf{22.0}/30.0 \\
 & UR5-WSG & 70.0/4.0 & \textbf{44.0}/8.0 & 82.5/20.0 & 80.0/22.0 & 80.0/24.0 & 82.0/10.0 & 78.0/\textbf{34.0} \\
 & Aloha-AgileX & 76.0/6.0 & 54.0/16.0 & 62.5/14.0 & 40.0/\textbf{44.0} & 28.0/22.0 & \textbf{18.0}/10.0 & 30.0/12.0 \\
 & \textit{Task avg.} & 69.2/4.0 & \textbf{49.2}/10.4 & 68.4/15.6 & 53.6/\textbf{32.4} & 57.2/21.2 & 52.0/10.8 & 50.4/27.6 \\
\cmidrule(lr){1-9}
Stack Two Blocks & Piper & 54.0/60.0 & \textbf{42.0}/20.0 & 55.0/61.0 & 52.0/60.0 & 54.0/\textbf{66.0} & 52.0/64.0 & 60.0/\textbf{66.0} \\
 & Franka-Panda & 32.0/90.0 & \textbf{34.0}/78.0 & 36.0/\textbf{86.0} & 76.0/80.0 & 40.0/74.0 & 36.0/82.0 & \textbf{34.0}/84.0 \\
 & ARX-X5 & 82.0/72.0 & 52.0/50.0 & 61.0/68.0 & 30.0/\textbf{80.0} & \textbf{26.0}/58.0 & 52.0/40.0 & 80.0/28.0 \\
 & UR5-WSG & 24.0/84.0 & 44.0/50.0 & 33.0/69.0 & 42.0/72.0 & 34.0/\textbf{74.0} & 24.0/56.0 & \textbf{20.0}/72.0 \\
 & Aloha-AgileX & 80.0/44.0 & 82.0/0.0 & \textbf{33.5}/0.0 & 60.0/\textbf{4.0} & 78.0/2.0 & 68.0/0.0 & 64.0/0.0 \\
 & \textit{Task avg.} & 54.4/70.0 & 50.8/39.6 & \textbf{43.7}/56.8 & 52.0/\textbf{59.2} & 46.4/54.8 & 46.4/48.4 & 51.6/50.0 \\
\cmidrule(lr){1-9}
Place Burger \& Fries & Piper & 84.0/18.0 & \textbf{26.0}/0.0 & 78.5/23.5 & 80.0/16.0 & 84.0/\textbf{30.0} & 84.0/24.0 & 82.0/28.0 \\
 & Franka-Panda & 100.0/6.0 & 100.0/4.0 & 100.0/1.5 & 100.0/\textbf{10.0} & 100.0/4.0 & 100.0/4.0 & \textbf{96.0}/2.0 \\
 & ARX-X5 & 100.0/0.0 & 74.0/2.0 & 84.0/9.0 & 30.0/\textbf{44.0} & 44.0/10.0 & \textbf{28.0}/8.0 & 42.0/12.0 \\
 & UR5-WSG & 100.0/2.0 & 90.0/2.0 & 92.5/1.5 & 84.0/12.0 & 84.0/\textbf{18.0} & 84.0/0.0 & \textbf{56.0}/0.0 \\
 & Aloha-AgileX & 70.0/6.0 & 84.0/\textbf{18.0} & \textbf{43.5}/1.0 & 48.0/6.0 & 54.0/14.0 & 56.0/0.0 & 58.0/0.0 \\
 & \textit{Task avg.} & 90.8/6.4 & 74.8/5.2 & 79.7/7.3 & 68.4/\textbf{17.6} & 73.2/15.2 & 70.4/7.2 & \textbf{66.8}/8.4 \\
\cmidrule(lr){1-9}
Stack Two Bowls & Piper & 8.0/28.0 & \textbf{6.0}/2.0 & 11.5/27.5 & 8.0/\textbf{36.0} & 8.0/26.0 & 10.0/30.0 & 12.0/30.0 \\
 & Franka-Panda & 14.0/88.0 & 10.0/62.0 & 16.0/79.5 & 22.0/70.0 & 60.0/40.0 & 20.0/\textbf{80.0} & \textbf{6.0}/76.0 \\
 & ARX-X5 & 30.0/66.0 & 50.0/10.0 & 27.5/43.5 & 12.0/\textbf{50.0} & \textbf{8.0}/48.0 & 10.0/40.0 & 14.0/36.0 \\
 & UR5-WSG & 14.0/86.0 & 34.0/24.0 & 29.5/69.5 & 12.0/82.0 & 24.0/76.0 & 22.0/48.0 & \textbf{6.0}/\textbf{88.0} \\
 & Aloha-AgileX & 30.0/48.0 & 26.0/0.0 & 10.0/0.5 & 14.0/\textbf{4.0} & 20.0/0.0 & \textbf{4.0}/0.0 & \textbf{4.0}/\textbf{4.0} \\
 & \textit{Task avg.} & 19.2/63.2 & 25.2/19.6 & 18.9/44.1 & 13.6/\textbf{48.4} & 24.0/38.0 & 13.2/39.6 & \textbf{8.4}/46.8 \\
\midrule
\multicolumn{2}{l}{\textbf{Overall average}} & 64.1/32.2 & 53.8/16.7 & 54.4/27.0 & 51.5/\textbf{35.0} & 53.4/30.0 & \textbf{49.7}/23.3 & 49.8/29.4 \\
\bottomrule
\end{tabular}
\end{table*}

\end{document}